\documentclass{article}

\usepackage{arxiv}

\usepackage[utf8]{inputenc} 
\usepackage[T1]{fontenc}    
\usepackage{hyperref}       
\usepackage{url}            
\usepackage{booktabs}       
\usepackage{amsfonts}       
\usepackage{nicefrac}       
\usepackage{microtype}      
\usepackage{lipsum}		
\usepackage{graphicx}
\usepackage[sort, numbers]{natbib}
\usepackage{doi}
\usepackage{multirow}
\usepackage{multicol}
\usepackage{array}
\usepackage{amsmath}
\usepackage[ruled,vlined]{algorithm2e}
\usepackage{subfigure}

\title{3DViT-GAT: A Unified Atlas-Based 3D Vision
Transformer and Graph Learning Framework for
Major Depressive Disorder Detection Using
Structural MRI Data}

\author{ \href{https://orcid.org/0000-0000-0000-0000}{\includegraphics[scale=0.06]{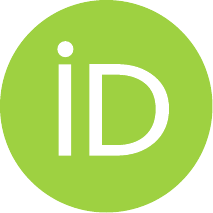}\hspace{1mm}Nojod M. Alotaibi*}\\
        Department of Computer Science, \\
        Faculty of Computing and Information Technology,\\
        King Abdulaziz University, Jeddah, Saudi Arabia\\
        \texttt{nalotaibi0351@stu.kau.edu.sa}\\
	\And
	\href{https://orcid.org/0000-0000-0000-0000}{\includegraphics[scale=0.06]{orcid.pdf}\hspace{1mm}Areej M. Alhothali*}\\
        Department of Computer Science, \\
        Faculty of Computing and Information Technology,\\
        King Abdulaziz University, Jeddah, Saudi Arabia\\
        \texttt{aalhothali@kau.edu.sa}\\
    \And
	\href{https://orcid.org/0000-0000-0000-0000}{\includegraphics[scale=0.06]{orcid.pdf}\hspace{1mm} Manar S. Ali} \\
	Department of Computer Science, \\
        Faculty of Computing and Information Technology,\\
        King Abdulaziz University, Jeddah, Saudi Arabia\\
}

\renewcommand{\shorttitle}{\textit{arXiv} Template}

\hypersetup{
pdftitle={A template for the arxiv style},
pdfsubject={q-bio.NC, q-bio.QM},
pdfauthor={David S.~Hippocampus, Elias D.~Striatum},
pdfkeywords={First keyword, Second keyword, More},
}

\begin{document}
\maketitle

\begin{abstract}
Major depressive disorder (MDD) is a prevalent mental health condition that negatively impacts both individual well-being and global public health. Automated detection of MDD using structural magnetic resonance imaging (sMRI) and deep learning (DL) methods holds increasing promise for improving diagnostic accuracy and enabling early intervention. Most existing methods employ either voxel-level features or handcrafted regional representations built from predefined brain atlases, limiting their ability to capture complex brain patterns. This paper develops a unified pipeline that utilizes Vision Transformers (ViTs) for extracting 3D region embeddings from sMRI data and Graph Neural Network (GNN) for classification. We explore two strategies for defining regions: (1) an atlas-based approach using predefined structural and functional brain atlases, and (2) an cube-based method by which ViTs are trained directly to identify regions from uniformly extracted 3D patches. Further, cosine similarity graphs are generated to model interregional relationships, and guide GNN-based classification. Extensive experiments were conducted using the REST-meta-MDD dataset to demonstrate the effectiveness of our model. With stratified 10-fold cross-validation, the best model obtained 81.51\% accuracy, 85.94\% sensitivity, 76.36\% specificity, 80.88\% precision, and 83.33\% F1-score. Further, atlas-based models consistently outperformed the cube-based approach, highlighting the importance of using domain-specific anatomical priors for MDD detection.
\end{abstract}


\section{Introduction}
\label{sec:introduction}
Major Depressive Disorder (MDD) is a common and extremely debilitating mental health condition that causes persistent low mood, cognitive impairment, and loss of interest in daily activities \cite{cui2024major}. According to the World Health Organization (WHO), more than 322 million people worldwide suffer from MDD, making it one of the leading causes of disability \cite{ world2017global}. Thus, it is crucial to diagnose MDD in an early and accurate manner to improve treatment outcomes and mitigate its negative economic and societal effects.

The conventional method of diagnosing MDD relies on clinical observations and patient-reported symptoms, which are susceptible to bias and inconsistency \cite{zhuo2019rise}. Therefore, leveraging neuroimaging techniques, particularly structural magnetic resonance imaging (sMRI), has gained increasing interest as a means of diagnosing MDD. sMRI provides high-resolution anatomical brain information, permitting researchers to identify structural abnormalities associated with the disorder, such as reduced gray matter (GM) volume in specific brain regions \cite{dai2019brain}.

Recently, deep learning (DL) has been widely recognized as one of the most effective methods for automating mental disorders diagnosis based on neuroimaging data. Specifically, Convolutional Neural Networks (CNNs) have shown great promise for MDD detection due to their ability to extract hierarchical spatial features from 3D brain volumes \cite{wang2021major}, \cite{hong20223d}. However, CNNs have several limitations, including their reliance on local receptive fields for feature extraction and their difficulty in modeling long-range spatial dependencies required for accurate diagnosis \cite{jia2025multimodal}. Furthermore, they exhibit limited generalizability, as their performance is often compromised when applied to datasets collected from multiple sites \cite{ takahashi2024comparison}.

To address these shortcomings, Vision Transformers (ViTs) have recently received significant attention for neuroimaging applications. Unlike CNNs, ViTs utilize self-attention mechanisms to capture local and global spatial dependencies, making them particularly useful for brain disorder classification tasks \cite{ takahashi2024comparison}. Several sMRI studies have demonstrated the effectiveness of ViT over CNN. For instance, Carcagnì et al. \cite{carcagni2023convolution}, Sarraf et al. \cite{sarraf2023ovitad}, and Shaffi et al. \cite{shaffi2024ensemble} found that ViTs performed better than CNNs at detecting Alzheimer’s disease. Additionally, Bi et al.\cite{bi2023multivit}, \cite{bi2024gray} have reported improved diagnostic accuracy with ViTs over CNN for diagnosis of schizophrenia. Although these findings are promising, few studies have examined ViTs for MDD classification using sMRI. However, most of these studies utilize patch-based methods, where sMRI volumes are uniformly divided into fixed-size 2D or 3D patches, thereby ignoring anatomically meaningful brain regions. To our knowledge, no prior work has systematically compared cube-based (3D patches) with atlas-based extraction strategies, where patches are derived from predefined brain atlases in ViT-based MDD detection frameworks.

This study addresses these gaps by proposing 3DViT-GAT, a unified atlas-based framework that integrates a 3D ViT and Graph Attention Network (GAT) for MDD detection utilizing sMRI data from the REST-meta-MDD dataset. Our key contributions are as follows:
\begin{enumerate}

    \item Region-based embedding extraction: We developed an atlas-based ViT pipeline that creates patches based upon meaningful regions of interest (ROIs). As a comparison, we also implemented a cube-based strategy using uniform 3D patches. For both pipelines, 3D ViT is used to extract region-level embeddings from sMRI data.
    \item Graph construction from ViT embeddings: Cosine similarity and K-nearest neighbors (KNN) are used to build graphs from ViT-derived embeddings, where nodes represent regions and edges indicate similarities between regions.
    \item Graph neural network classification: The resulting graphs are used to train a GAT model for MDD classification, which effectively models both intra- and inter-regional dependency.
    \item Comprehensive evaluation: We performed stratified 10-fold cross-validation to compare the atlas-based (our proposed model) and cube-based pipelines, showing that 3DViT-GAT performed better than ViT pipelines utilizing a cube-based extraction method.
\end{enumerate}

The rest of the paper is organized as follows. Section \ref{sec:related work} provides a brief overview of the most relevant studies. A brief description of the dataset and its preprocessing procedure is provided in \ref{sec:dataset}. Section \ref{sec: methodology} introduces the proposed methodology, which includes region extraction strategies, ViT and GAT methods. Section \ref{sec:results} outlines experimental results and evaluation methods, followed by a discussion of model performance compared with baseline methods.
Finally, the conclusion of this paper is summarized in \ref{sec:conclusion}.

\section{RELATED WORK}
\label{sec:related work}
Structural MRI (sMRI) has been increasingly employed for the automated detection of major depressive disorder (MDD) using both machine learning (ML) and deep learning (DL) techniques.

Early ML-based methods relied on handcrafted feature extraction followed by classical classifiers. For instance, Kim et al. \cite{kim2019machine} proposed a highly accurate prediction model to distinguish adolescent MDD patients from their healthy counterparts (HC). As part of the classification task, the features were extracted using unpaired t-tests and then fed into many ML classifiers, including random forest (RF) and support vector machines (SVM). Finally, the leave-one-out cross-validation method (LOOCV) was used to evaluate the proposed model and the best performance was achieved by the SVM classifier with 94.4\% accuracy. Ma et al. \cite{ma2022gray} introduced a radiomic-based approach to diagnose MDD, subthreshold depressions (StD), and HC. Structural radiomic features were extracted from sMRI and diffusion tensor imaging (DTI) of gray matter (GM) and white matter (WM). The RF algorithm embedded in 10-fold cross validation was used to remove redundant and non-relevant features. The proposed model achieved accuracies of 86.75\% for MDD vs. HC, 70.51\% for STD vs. HC, and 54.55\% for MDD vs. STD.

Later, DL-based approaches became prominent due to their ability to automatically learn discriminative features. Mousavian et al. \cite{mousavian2019depression} proposed a framework where raw features were first obtained by applying entropy-based slice selection to sMRI images. After this, the data was augmented by either duplication or rotation in order to create a balanced dataset. Subsequently, DL models, including convolutional neural network (CNN) and pretrained networks such as VGG16 and Inceptionv3 were employed to extract additional feature representations. Finally, several classifiers, including non-linear kernel SVM (RBF) and linear kernel SVM, were then trained separately on the raw features and the extracted features. Their results demonstrated that the highest accuracy, reaching 96\%, was achieved by training an SVM-RBF classifier directly on the raw features. 

More recent work has explored 3D CNN for MDD classification using sMRI data and utilizing their capacity to capture spatial hierarchy within brain data. Wang et al. \cite{wang2021major} developed a voxel-wise densely connected convolutional neural network known as 3D-DenseNet to classify and predict MDD using sMRI. To improve model performance, they introduced a novel ADNI-Transfer strategy, where the model was pretrained on the Alzheimer's Disease Neuroimaging Initiative (ADNI) dataset. Their experiments demonstrated that the 3D-DenseNet model with ADNI-Transfer achieved a classification accuracy of 84.37\% for MDD diagnosis, significantly outperforming other ML and DL approaches. Similarly, Hong et al. \cite{hong20223d} proposed an automated 3D FRN-ResNet framework for the identification of MDD from sMRI images. This approach combined a 3D-ResNet for volumetric feature extraction with a Feature Map Reconstruction Network (FRN) designed to preserve critical spatial and location-information for the extracted features. Further, their model was evaluated using the dataset from the Seventh Hospital of Hangzhou (SHH). The experimental results showed that the 3D FRN-ResNet model achieved superior performance compared to other traditional methods, reaching a classification accuracy of 86.776\%.

Recently, several studies have included attention mechanisms into DL frameworks in an effort to enhance both interpretability and classification performance for sMRI-based MDD detection. Among them, Gao et al. \cite{gao2023classification} developed an attention-guided framework for MDD classification using sMRI data, integrating both voxel-based and region-based features in a two-stage process. Initially, a modified 2D BrainNetCNN was applied to individual structural covariance networks constructed using the Automated Anatomical Labeling (AAL) atlas to perform primary classification and produce a heatmap highlighting connections of positive influence. This heatmap was then utilized to enhance the original 3D gray matter (GM) images by emphasizing significant regions. Afterwards, a lightweight 3D CNN was trained on the enhanced GM images for the final classification task. The results indicated that the proposed model obtained a classification accuracy of 76.65\% on the REST-meta-MDD dataset. Xiao et al. \cite{10925117} proposed a novel deep learning framework named SQUID, which is an anomaly detection framework for diagnosing MDD using sMRI GM slices. This model divides sMRI images into non-overlapping blocks and encodes them using an Auto Encoder (AE) to extract key features. Afterwards, the images are reconstructed based on the extracted features using a Memory Queue image reconstruction module guided by a Transformer. Lastly, the discriminator evaluates the reconstructed images to distinguish between MDD patients and HCs. The results showed that the SQUID model delivered a classification accuracy of 77.17\% using the REST-meta-MDD dataset.

However, most prior work on sMRI-based MDD classification relied on handcrafted features combined with ML models, or on DL trained on raw brain images or fixed-size image patches. Often, these approaches fail to capture the subtle spatial relationships between brain regions, which are essential for understanding the complex patterns of MDD. Moreover, handcrafted methods are highly dependent on the quality of manually extracted features, which may result in limited generalizability and robustness. The absence of anatomical guidance in patching strategies can produce a loss of important structural information, thereby reducing the accuracy and interpretability of classification models. It is noteworthy that no prior studies have systematically compared anatomically-informed (atlas-based) with uninformed (cube-based) region extraction strategies for MDD detection.

To address these gaps, we proposed a novel 3DViT-GAT framework for MDD detection based on sMRI data. In this architecture, each brain volume is first divided into regions of interest (ROIs) based on anatomical atlases, enabling the model to focus on semantically meaningful regions.  A 3D ViT model is employed to extract embeddings from each ROI, which capture local and global spatial patterns. These embeddings are then used to create subject-wise graphs, in which nodes represent ROIs and edges reflect top-k cosine similarity connections. Finally, a GAT model is trained to analyze these graphs by highlighting the most informative ROI connections, improving the accuracy of MDD classification. As a further evaluation of the influence of anatomical priors, we compared our atlas-based ViT-GNN model with a cube-based variant where regions are extracted via fixed-size 3D patches.

\section{DATASET DESCRIPTION}
\label{sec:dataset}
\subsubsection{REST-meta-MDD DATASET}
This study utilizes a dataset from the REST-meta-MDD consortium, which to our knowledge is the largest public MDD dataset \cite{chen2022direct}. This dataset comprises 2,428 participants from 25 sites, including 1,300 MDD patients (826 females and 474 males) and 1,128 HC \cite{chen2022direct}, \cite{yan2019reduced}. Moreover, there are two types of imaging data, T1-weighted sMRI and rs-fMRI, collected at each site along with phenotypic data such as age, gender, medication status, episode status (recurrent or first episode), illness duration, and the 17-item Hamilton Depression Rating Scale (HAMD) \cite{chen2022direct}. REST-meta-MDD participants provided written informed consent prior to participating, and the local Institutional Review Board approved the collection of data at each site \cite{chen2022direct}, \cite{yan2019reduced}.
\subsubsection{DATA PREPROCESSING}

The T1-weighted sMRI scans were preprocessed using Diffeomorphic Anatomical Registration Through Exponentiated Lie algebra (DPARSF) software \cite{yan2019reduced}. Preprocessing involved discarding the first ten volumes to ensure magnetization equilibrium, correcting head motion, segmenting into gray matter (GM), white matter (WM), and cerebrospinal fluid (CF), normalizing to the Montreal Neurological Institute (MNI) template, and spatial smoothing. Additionally, a duplicate site was discovered after consortium data sharing and discarded, and covariates were regressed to ensure data quality and consistency \cite{yan2019reduced}. The resulting sMRI dataset contained 1586 subjects, of which 1276 were  MDD patients and 1104 were HC.

Further, we utilized GM volumes in the MNI space to represent the original sMRI data. Each sMRI image had a standardized spatial dimension of $121 \times 145 \times 121$ voxels. A detailed description of the demographic characteristics of the subjects included in our study is provided in Table \ref{tab:DemoInfo}.
.

\begin{table}[h]
\centering
\caption{Demographic Information of the 1563 Study Subjects.}
\label{tab:DemoInfo}
    \renewcommand{\arraystretch}{1.5}
    \begin{tabular}{|c|c|c|c|c|c|} \hline 
    \multirow{2}{*}{\textbf{Group}} & 
    \multirow{2}{*}{\textbf{Number of subjects}}& 
    \multirow{2}{*}{\textbf{Male}}& 
    \multirow{2}{*}{\textbf{Female}} &
    \multicolumn{2}{c|}{\textbf{Average $\pm$ Standard deviation (\%)} }\\ 
    \cline{5-6}
    &&&&
    \textbf{Age} & \textbf{Education}\\
    \hline 
    \begin{tabular}{c} MDD  \end{tabular}& 
    \begin{tabular}{c}1276\end{tabular}&  \begin{tabular}{c}463\end{tabular}&  \begin{tabular}{c}813\end{tabular}& \begin{tabular}{c}36.23$\pm$14.62\end{tabular}& \begin{tabular}{c}11.25$\pm$4.20\end{tabular}\\ \hline
    \begin{tabular}{c}HC  \end{tabular}&  \begin{tabular}{c}1104\end{tabular}&  \begin{tabular}{c}462\end{tabular}&  \begin{tabular}{c}642\end{tabular}& \begin{tabular}{c}36.15$\pm$15.66\end{tabular}& \begin{tabular}{c}12.25$\pm$4.98\end{tabular}\\ \hline
\end{tabular}
\end{table}

\section{METHODOLOGY}
\label{sec: methodology}
\begin{figure}
    \centering
    \includegraphics[width=\textwidth,height=0.70\textheight]{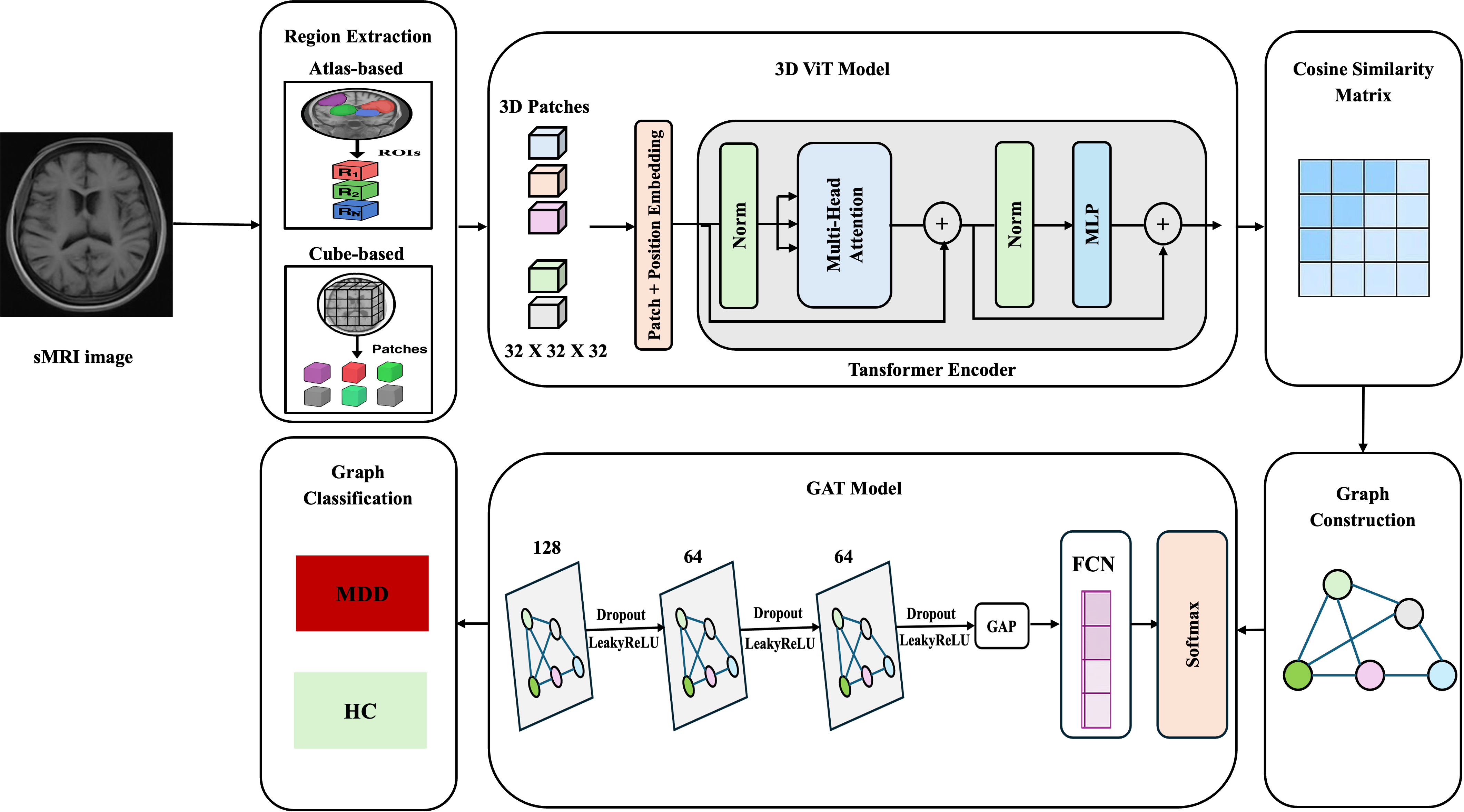}
    \caption{An overview of the proposed 3DViT-GAT model. sMRI, structural MRI; ViT, vision transformer; MLP, multilayer perceptron; GAT, graph attention network; LeakyRelu, leaky rectified linear unit; GAP, global average pooling; FCN, fully connected network; MDD, major depressive disorder; HC, healthy control; ROI, brain regions of interest; N, the number of ROIs.}
    \label{fig:3DViT-GAT}
\end{figure}

This study presents a comprehensive framework for detecting MDD using sMRI data from the REST-meta-MDD dataset. This framework incorporates a 3D Vision Transformer (ViT) to extract brain region features and a graph neural network (GNN) to perform graph-level classification. The overall architecture of the proposed method is illustrated in Figure \ref{fig:3DViT-GAT}. First, we extract region embeddings from each 3D sMRI image by using two different strategies: atlas-based and cube-based. In the atlas-based approach, the brain is segmented into semantically meaningful regions (ROIs) using four popular brain atlases, including Automated Anatomical Labeling (AAL) atlas \cite{tzourio2002automated}, Harvard–Oxford (HO) atlas \cite{kennedy1998gyri}, Dosenbach’s 160 functional ROIs (Dose) \cite{dosenbach2010prediction}, and Craddock’s clustering 200 ROIs (CK) \cite{craddock2012whole}. In contrast, the cube-based approach divides the whole brain into uniform 3D cubes without using any brain atlas. The extracted 3D ROIs or cubes are then treated as patches and fed separately into a 3D ViT model to determine discriminative embeddings. After that, we construct a cosine similarity matrix for each subject to model the inter-regional relationship between these embeddings. The similarity matrices are then represented as graphs. Finally, a Graph Attention Network (GAT) is applied to classify subjects based on these graphs, performing a graph classification task. 

\subsection{REGION EXTRACTION STRATEGIES}

\subsubsection{ATLAS-BASED EXTRACTION}
Through this strategy, the brain is divided into meaningful regions based on predefined atlases. Specifically, this study used two structural atlases, AAL (116 ROIs) and HO (112 ROIs), as well as two functional atlases, Dose (160 ROIs) and CK (200 ROIs). Atlases provide a consistent spatial template that defines ROI boundaries, which facilitates the extraction of regions from sMRI scans.

In this case, each ROI is extracted as a volumetric patch, whereby the number of patches for each subject is equal to the number of ROIs defined by the selected atlas. The patch size is fixed to $32 \times 32 \times 32$ voxels to ensure uniform input dimensions across subjects and regions. These 3D patches represent localized structural features and serve as inputs to the 3D ViT model. Consequently, the model can learn region-specific discriminative representations, enhancing performance and accuracy for a subsequent graph classification task. 
\subsubsection{CUBE-BASED EXTRACTION}
In this method, the original sMRI images of size $121 \times 145 \times 121$ voxels are resized to a standardized shape of $120 \times 140 \times 120$ voxels. The resized images are then uniformly partitioned into non-overlapping 3D cubes of size $32 \times 32 \times 32$ voxels. This cube size was chosen to ensure a balance between spatial resolution and computational efficiency. The total number of extracted cubes (patches) per subject is calculated as follows:
\begin{equation}
\text{Number of patches} = \left\lfloor\frac{120}{32} \right\rfloor\times
\left\lfloor\frac{140}{32}\right\rfloor\times
\left\lfloor\frac{120}{32}\right\rfloor = 36
\label{eq:patches}
\end{equation}

The fixed number of patches per subject maintains consistency across the dataset and provides a standardized input format for downstream processing. Afterwards, the extracted 3D patches are used as input for the 3D ViT model to capture region-based discriminative representations for each individual.
\subsection{VISION TRANSFORMER MODEL}
The Vision Transformer (ViT) is a recent DL architecture that has exhibited impressive performance across various computer vision tasks, primarily image classification \cite{dosovitskiy2020image}. Contrary to CNN, the ViT is capable of modeling local and global dependencies across image patches through self-attention, allowing it to focus on features more relevant to classification \cite{dai2024classification}.

As part of this study, we employed a 3D ViT model to process sMRI images for the MDD classification task. The 3D ViT model operates using uniformly sized 3D patches, which are either extracted by a predefined atlas or by a non-overlapping partitioning process. In this manner, the model can leverage the spatial context to detect subtle differences in sMRI images indicative of MDD, increasing both the accuracy and robustness of classification.
\subsubsection{MODEL ARCHITECTURE}
The 3D ViT model requires the input of a sequence of 3D patches: $x_{p} = [x_{p}^{1}, x_{p}^{2}, \dots, x_{p}^{N}]$. Each patch of size $x_{i}\in R^{(32 \times 32 \times 32)}$ is flattened and reshaped as a vector and then projected into a lower-dimensional embedding space using a linear patch embedding layer \cite{dosovitskiy2020image}:
\begin{equation}
z_0^i = E x_{p}^{i}, \quad i = 1, \dots, N  
\end{equation}

Here, $E \in R^{(32^{3})}$ is the linear projection matrix, and $z^{i}_{0}$ is the patch embedding. In addition, a patch learnable classification token $z^{cls}_{0}$ is also prepended to the sequence as follows \cite{dosovitskiy2020image}:
\begin{equation}
Z_{0} = [z_{0}^{\text{cls}}; z_{0}^{1}; z_{0}^{2}; \dots; z_{0}^{N}] + E_{pos}
\end{equation}

Where $E_{pos}$ is the learned positional encoding matrix for the full input sequence. This sequence is then passed through a stack of Transformer encoder layers (L = 6), which include multiheaded self-attention (MHSA) and multilayer perception (MLP) blocks. The LayerNorm (LN) is applied before each block, and the residual connections are applied after each block as follows:
\begin{equation}
Z_{l}^{\prime} = \text{MHSA}\left(\text{LN}\left(Z_{l-1}\right)\right) + Z_{l-1} 
\end{equation}
\begin{equation}
Z_{l} = \text{MLP}\left(\text{LN}\left(Z_l\right)\right) + Z_l
\end{equation}

The mechanism of self-attention operation is described as follows \cite{dosovitskiy2020image}:
\begin{equation}
    \text{Attention}(Q, K, V) = \text{softmax}\left(\frac{QK^T}{\sqrt{d_k}}\right)V
\end{equation}

In this formula, $Q, K,$ and $V$ represent query, key, and value matrices, respectively, which are computed from input embeddings using learned linear projections. The dimensionality of each attention head $d_{k}$ is calculated by dividing the embedding dimension $d$ by the number of attention heads $h$. In our configuration, we set the embedding dimension to 128 and the number of attention heads to 8, resulting in $d_k = \frac{128}{8} = 16$. As a result, the model can simultaneously capture diverse contextual information from multiple spatial perspectives.

Ultimately, the model is fine-tuned based on the classification objective to maximize its performance. Specifically, all patch embeddings are concatenated with the cls embedding to create the classification head input, rather than simply using the cls token. 
\begin{equation}
Z_{\text{final}} = [z_L^{\text{cls}} \, \| \, z_L^1 \, \| \, z_L^2 \, \| \,\dots \, \| \, z_L^N]
\end{equation}

This concatenation preserves both global and regional features, which enhances the model's discriminative power. This enriched representation $Z_{\text{final}}$ is then fed into a fully connected classification layer (FCN) with softmax activation to produce the final prediction: 

\begin{equation}
    \hat{y} = \text{softmax}(W Z_{\text{final}} + b)
\end{equation}

Where $W$ and $b$ are the weight and bias of the final classification layer, and $\hat{y}$ is the predicted class probability.

The detailed configuration of the 3D ViT used in our experiments can be found in Table \ref{tab:vit_params}. We trained the model using the adaptive Moment Estimation (Adam) optimizer and cross-entropy loss regularized using a L2 regularization technique called weight decay. In addition, a stratified 10-fold cross-validation process was employed during the training process to ensure robust generalization.
\begin{table}[h]
\caption{3D ViT Model Configuration and Training Hyperparameters}
\centering
\renewcommand{\arraystretch}{1}

\begin{tabular}{|c|c|}

\hline

\textbf{Parameter} & \textbf{Value} \\

\hline
Patch size & $32 \times 32 \times 32$ \\
\hline
Embedding dimension & 128 \\
\hline
Number of transformer layers & 6 \\
\hline
Number of attention heads & 8 \\
\hline
Learning rate & $1 \times 10^{-5}$ \\
\hline
Weight decay & $1 \times 10^{-8}$ \\
\hline
Dropout & 0.3 \\
\hline
Batch size & 16 \\
\hline
Optimizer & Adam \\
\hline
Loss function & Cross-entropy \\
\hline
Classification head & Fully connected layer \\
\hline

\end{tabular}

\label{tab:vit_params}
\end{table}
\subsection{GRAPH CONSTRUCTION}
To capture the inter-regional relationships between brain regions, we constructed a subject-specific graph based on region-wise embeddings obtained from the 3D ViT model. In each graph, nodes represent individual regions defined as anatomical ROIs (in the atlas-based model) or as fixed-size 3D cubes (in the cube-based model). The edges between these nodes are determined by computing the cosine similarity between each node embedding. The cosine similarity between two embeddings $z_{i}$ and $z_{j}$ can be calculated using the following formula \cite{bi2024gray}:
\begin{equation}
    \text{Cosine\_similarity}(z_i, z_j) = \frac{z_i \cdot z_j}{\|z_i\| \|z_j\|}
\end{equation}

In this case, the numerator $z_i \cdot z_j$ refers to the product of $z_{i}$ and $z_{j}$, while the denominator $\|z_i\| \|z_j\|$ refers to the product of the Euclidean norms of these embeddings. The cosine similarity score ranges from -1 to 1, where 1 indicates complete similarity, 0 indicates orthogonality, and -1 indicates dissimilarity.

This similarity matrix of size $N \times N$, where $N$ is the number of regions is converted into a graph structure by applying a K-nearest neighbors (KNN) strategy with $K = 10$, which keeps only the top-K similar neighbors for each node, denoted as $\mathcal{N}_K(v_i)$. Afterward, the undirected edges are added from $v_i$ to its top-K neighbors. Furthermore, edge weights are set according to their respective similarity values $S_{ij}$, resulting in a weighted undirected graph $G$. This graph construction assists in learning spatial relationships among brain regions, which is vital to downstream graph classification tasks. An illustration of the entire graph construction process is provided in Algorithm \ref{alg:cosine_knn_graph}.

\begin{algorithm}
\caption{Graph Construction Process}
\label{alg:cosine_knn_graph}
\KwIn{Set of ViT embeddings $Z = \{z_1, z_2, \dots, z_N\}$, Set of nodes $V = \{v_1, v_2, \dots, v_N\}$ , number of neighbors $K=10$} 
\KwOut{Subject-specific graph $G = (V, E)$, where $E$ is the set of edges}
\vspace{0.5em}
\textbf{Graph Construction:\\} 
\For{each node $v_i \in V$}{
\textbf{1.} \For{each node $v_j \in V$}{
Compute the cosine similarity matrix $S \in {R}^{N \times N}$ such that: $S_{ij}=cosine\_similarity(z_i,z_j)$ }
\textbf{2.} Identify the set $\mathcal{N}_K(v_i)$ of $K$ most similar nodes to a node $v_i$ based on $S$ \\  \textbf{3.} \For{each node $v_h \in \mathcal{N}_K(v_i)$, $h \neq i$}{Add undirected edge $(v_i, v_h)$ to the graph $G$}
}
\vspace{0.5em}
\Return {$G$} 
\end{algorithm}

\subsection{GRAPH ATTENTION NETWORK (GAT)}
In graph-based learning, a Graph Attention Network (GAT) is a type of Graph Neural Network (GNN) that utilizes an attention mechanism to allocate different weights to neighboring nodes according to their relevance to the target node \cite{velivckovic2017graph}. In this manner, GAT can detect complex relationships and dependencies in graph-structured data, making them capture complex inter-regional relationships in brain graphs more effectively than traditional GNNs.

Formally, consider a graph G = (V, E), where V is the set of nodes (ROIs or 3D cubes) and E is the set of edges (similarity relationships). Each node $v_i \in V$ is associated with a feature vector $\mathbf{z}_i \in \mathbb{R}^F$ representing its ViT-derived embedding.  In the GAT, updated node embedding $\mathbf{z}_i^{'} \in \mathbb{R}^{F^{'}}$ is produced by aggregating information from its neighboring nodes $\mathcal {N}_{(v_i)}$ using attention coefficients $\alpha_{ij}$ across various attention heads. The updated representation of node $v_i$ is calculated as follows \cite{velivckovic2017graph}:
\begin{equation}
{z}_{i}^{'} = \parallel_{k=1}^{K} \sigma \left( \sum_{j \in \mathcal{N}_{(v_i)}} \alpha_{ij}^{k} \mathbf{W}^{k} \mathbf{z}_j \right)
\end{equation}

Where $\alpha_{ij}$ represents the normalized attention coefficient, $K$ is the number of attention heads, $W^k$ is a linear weight matrix for the $k$-th attention head, $\sigma$ is a nonlinear activation function, and $\parallel_{k=1}^{K}$ reflects the concatenation of outputs of all attention heads. The normalized attention coefficient $\alpha_{ij}$ is computed as:
\begin{equation}
\label{leaky}
\alpha_{ij}=\frac{exp(LeakyReLU({a}^T[W {z}_{i} \parallel W {z}_{j}]))}{\sum_{h \in \mathcal{N}_{(v_i)}} exp(LeakyReLU({a}^T[W {z}_{i}\parallel W {z}_{h}]))}
\end{equation}

Here, $a$ is a learnable attention vector shared across all nodes, the Leaky Rectified Linear Unit (LeakyReLU) is a nonlinear function, ${.}^{T}$ refers to a transposition, and $\parallel$ is the concatenation operation.
\subsubsection{MODEL ARCHITECTURE}
Our GAT model comprises three stacked layers of GATConv combined with LeakyReLU activation functions for learning the representation of each node. Each GATConv layer contains 64 hidden units. The number of attention heads used in this study is 4. To prevent overfitting, the dropout technique is implemented before each GATConv layer. In addition, the global average pooling (GAP) function is applied to obtain a representation of the entire graph by taking the average of all node embeddings as indicated in \cite{Cui_2023}:
\begin{equation}\label{GAPEq}
    z_{G} = \frac{1}{N} \sum_{i=1}^{N} {z}_i
\end{equation}

Assume that $z_{G}$ is the representation of a single graph $G$, $N$ is the number of nodes in $G$, and $z_i$ is the embedding of node $v_i$ in $G$. Finally, the graph representations are passed through a fully connected layer (FCN) and a Softmax activation to produce class probabilities.

The model is trained using the Adam optimizer with a learning rate of $5 \times 10^{-4}$ and a weight decay of $3 \times 10^{-5}$. A stratified 10-fold cross-validation technique was used to ensure a robust and unbiased evaluation. Table \ref{tab:GAT_params} summarizes the full hyperparameter configuration.

\begin{table}[htbp]
\caption{GAT Model Configuration and Training Hyperparameters}
\centering
\renewcommand{\arraystretch}{1}
\begin{tabular}{|c|c|}

\hline

\textbf{Parameter} & \textbf{Value} \\
\hline
Number of GAT layers & 3 \\
\hline
Hidden units per layer & 64 \\
\hline
Number of attention heads & 4 \\
\hline
Learning rate & $5 \times 10^{-4}$ \\
\hline
Weight decay & $3 \times 10^{-5}$ \\
\hline
Dropout & 0.3 \\
\hline
Activation function & LeakyRelu \\
\hline
Batch size & 16 \\
\hline
Optimizer & Adam \\
\hline
Loss function & Cross-entropy \\
\hline

\end{tabular}
\label{tab:GAT_params}
\end{table}

\section{RESULTS AND DISCUSSION}
In this section, we first present the evaluation metrics that we employed for the classification task. Then, we examine the prediction performance of our proposed model using a large-scale sMRI dataset derived from the REST-meta-MDD project.
\label{sec:results}
\subsection{PERFORMANCE EVALUATION}
Throughout this study, all experiments are conducted using PyTorch and training is carried out on the Aziz Supercomputer equipped with an NVIDIA A100 GPU, accelerating the training of deep learning models, especially when processing MRIs and graphs. As part of the evaluation process, we analyzed model performance using a set of standard evaluation metrics, such as accuracy, sensitivity, specificity, precision, and F1-score. These metrics are defined as follows \cite{SHARMA202231}:
\begin{equation} \label{eqAcc}
Accuracy =\frac{(TP + TN)}{(TP + TN + FP + FN)}
\end{equation}
\begin{equation} \label{eqSn}
Sensitivity = \frac{TP}{(TP + FN)} 
\end{equation}
\begin{equation} \label{eqSp}
Specificity = \frac{TN}{(FP + TN)} 
\end{equation}
\begin{equation} \label{eqP}
Precision = \frac{TP}{(TP + FP)}
\end{equation}
\begin{equation} \label{eqF1}
F1-score = \frac{(2 \times Precision \times Sensitivity)}{(Precision + Sensitivity)}
\end{equation}

Here, $TP$ indicates the correct classification of positive samples (MDD), $TN$ indicates the correct classification of negative samples (HC), $FP$ indicates the incorrect classification of negative samples as positive, and $FN$ indicates the incorrect negative classification.
\subsection{EXPERIMENTAL RESULTS}
In this section, we conduct an extensive analysis to evaluate the effectiveness of our proposed 3DViT-GAT framework under several experimental configurations. Three main scenarios are considered in our investigation. First, we assess the standalone performance of the 3D ViT model using two distinct input paradigms: (1) atlas-based extraction, which guides region segmentation using brain atlases, and (2) cube-based extraction, which divides the brain into uniform, non-overlapping 3D cubes. Second, our full 3DViT-GAT hybrid model is constructed by integrating ViT-derived embeddings into a GAT. Graph-based classification is similarly investigated utilizing both atlas-based and cube-based graph representations. Third, we compare the performance of our model with some of the state-of-the-art methods to demonstrate its competitiveness and generalizability for detecting MDD based on sMRI data. Finally, a model explainability analysis are conducted to identify the most critical brain ROIs associated with MDD, HC, and MDD vs. HC. 
\subsubsection{RESULTS OF 3D VIT-BASED MODELS}
We evaluated the classification performance of the proposed 3D ViT model using two different extraction strategies: atlas-based extraction and cube-based extraction. In the atlas-based approach, the brain is segmented into anatomically defined regions (ROIs) using four brain atlases (AAL, HO, Dose, and CK). Conversely, the cube-based approach divides the brain into volumetric regions of uniform size. 

Table \ref{tab:vit_models_performance} summarizes the average classification performance of the 3D ViT-based models across 10-fold cross-validation. Among the tested models, the HO-based model achieved the highest accuracy of 76.93$\pm$0.83\%, along with the highest specificity (73.36$\pm$6.58\%) and precision (78.09$\pm$3.34\%), indicating its strength in correctly identifying negative cases (HC) and minimizing false positives. In addition, it achieved strong performance in terms of sensitivity (80.00$\pm$5.07\%) and F1-score  (78.82$\pm$1.11\%). The Cube-based model produced the second highest accuracy at 76.81$\pm$1.27\% despite the absence of anatomical priors. It also exhibited the highest sensitivity (83.67$\pm$6.28\%) and F1-score (79.45$\pm$1.39\%), reflecting its greater ability to detect true positive cases (MDD). However, this improvement came at the expense of increased false positives, as reflected in its lower specificity (68.82$\pm$8.26\%) and precision (76.16$\pm$3.59\%). The Dose-based model yielded 76.72$\pm$1.45\% accuracy, 82.19$\pm$3.69\% sensitivity, 70.36$\pm$6.14\% specificity, 76.58$\pm$3.08\% precision, and 79.15$\pm$1.05\% F1-score. Meanwhile, the AAL-based model attained 76.60$\pm$0.98\%	accuracy, 81.56$\pm$5.80\% sensitivity, 70.82$\pm$6.84\% specificity, 76.80$\pm$3.07\% precision, and 78.88$\pm$1.42\% F1-score. Furthermore, The CK-based model delivered comparable performance, with an accuracy of 76.39$\pm$1.40\%, a sensitivity of 80.55$\pm$3.23\%, a specificity of 71.55$\pm$5.93\%, a precision of 76.95$\pm$3.17\%, and an F1-score of 78.59$\pm$0.87\%.

Moreover, Table \ref{tab:Best_ViTFold} displays the results obtained by the fold that achieved the highest accuracy. Among the atlas-based models,the best Dose-based model among all 10 folds achieved 78.99\% accuracy, 78.91\% sensitivity, 79.09\% specificity, 81.45\% precision, and 80.16\% F1-score. The best CK-based model had 78.15\% accuracy, 81.25\% sensitivity, 74.55\% specificity, 78.79\% precision, and 80.00\% F1-score. In the AAL-based model, the best model produced 77.73\% accuracy, 92.97\% sensitivity, 60.00\% specificity, 73.01\% precision, and 81.79\% F1-score. Similarly, the best HO-based model yielded 77.73\% accuracy, 83.59\% sensitivity, 70.91\% specificity, 76.98\% precision, and 80.15\% F1-score. 
On the other hand, the best cube-based model had 78.57\% accuracy, 87.50\% sensitivity, 68.18\% specificity, 76.19\% precision, and 81.45\% F1-score.

In summary, the atlas-based models showed more stable and balanced performance across all evaluation metrics, making them more suitable for real-world diagnostic settings where balancing sensitivity and specificity is crucial. Overall, these findings provide evidence for the superior discriminatory power of atlas-based region extraction over cube-based partitioning, which uses anatomical brain partitioning to provide more informative and robust feature representations for ViT-based MDD classification.
 
\begin{table}[h]
\caption{Average Classification Results and Standard Deviation of 10-fold Cross-validation for 3D ViT-based Models.}
\label{tab:vit_models_performance}
\renewcommand{\arraystretch}{1.8}
  \begin{tabular}{|cc|c|c|c|c|c|}\hline 
    \multicolumn{2}{|c}{\multirow{2}{*}{\textbf{Method}}} & 
    \multicolumn{5}{|c|}{ \centering \textbf{Average} $\pm$ \textbf{Standard deviation (\%)} } \\
     \cline{3-7} 
     & & \centering \textbf{Accuracy} & \centering \textbf{Sensitivity} & \centering 
     \textbf{Specificity} & \centering \textbf{Precision} & \centering\arraybackslash 
     \textbf{F1-score} \\
     \hline 
     \multicolumn{1}{|c}{\multirow{4.3}{*}{ \rotatebox[origin=c]{90}{Atlas-based models}}}& 
     \multicolumn{1}{|c}{\centering AAL-based model} & 
     \multicolumn{1}{|c}{ \begin{tabular}{c} 76.60$\pm$0.98\end{tabular}} & 
     \multicolumn{1}{|c}{\begin{tabular}{c}81.56$\pm$5.80 \end{tabular}} & \multicolumn{1}{|c}{\begin{tabular}{c} 70.82$\pm$6.84\end{tabular}} & 
     \multicolumn{1}{|c}{\begin{tabular}{c} 76.80$\pm$3.07\end{tabular}} & 
     \multicolumn{1}{|c|}{\begin{tabular}{c} 78.88$\pm$1.42\end{tabular}}\\ 
     \cline{2-7} 
     & \multicolumn{1}{|c}{\centering HO-based model} & \multicolumn{1}{|c}{\begin{tabular}{c} \textbf{76.93$\pm$0.83}\end{tabular}} & \multicolumn{1}{|c}{\begin{tabular}{c}80.00$\pm$5.07 \end{tabular}} & \multicolumn{1}{|c}{\begin{tabular}{c}\textbf{73.36$\pm$6.58}\end{tabular}} & \multicolumn{1}{|c}{\begin{tabular}{c}\textbf{78.09$\pm$3.34}\end{tabular}} & \multicolumn{1}{|c|}{\begin{tabular}{c} 78.82$\pm$1.11\end{tabular}}\\
     \cline{2-7} 
     &  \multicolumn{1}{|c}{\centering Dose-based model} & \multicolumn{1}{|c}{\begin{tabular}{c}76.72$\pm$1.45\end{tabular}} & \multicolumn{1}{|c}{\begin{tabular}{c}82.19$\pm$3.69\end{tabular}} & \multicolumn{1}{|c}{\begin{tabular}{c}70.36$\pm$6.14\end{tabular}} & \multicolumn{1}{|c}{\begin{tabular}{c}76.58$\pm$3.08\end{tabular}} & \multicolumn{1}{|c|}{\begin{tabular}{c}79.15$\pm$1.05\end{tabular}}\\ 
     \cline{2-7} 
     & \multicolumn{1}{|c}{\centering CK-based model} & \multicolumn{1}{|c}{\begin{tabular}{c} 76.39$\pm$1.40\end{tabular}} & \multicolumn{1}{|c}{\begin{tabular}{c} 80.55$\pm$3.23\end{tabular}} & \multicolumn{1}{|c}{\begin{tabular}{c}71.55$\pm$5.93\end{tabular}} & \multicolumn{1}{|c}{\begin{tabular}{c}76.95$\pm$3.17\end{tabular}} & \multicolumn{1}{|c|}{\begin{tabular}{c} 78.59$\pm$0.87\end{tabular}}\\ 
     \hline 
     \multicolumn{2}{|c}{Cube-based model}
    & \multicolumn{1}{|c}{\begin{tabular}{c}76.81$\pm$1.27 \end{tabular}} & 
    \multicolumn{1}{|c}{\begin{tabular}{c}\textbf{83.67$\pm$6.28}\end{tabular}} & \multicolumn{1}{|c}{\begin{tabular}{c}68.82$\pm$8.26\end{tabular}} & 
    \multicolumn{1}{|c}{\begin{tabular}{c}76.16$\pm$3.59 \end{tabular}}& \multicolumn{1}{|c|}{\begin{tabular}{c}\textbf{79.45$\pm$1.39}\end{tabular}} \\
    \hline
     
  \end{tabular}
{\\\\The bolded values indicate the best performing results.}
\end{table}

\begin{table}[h]
\caption{Classification Results for the Fold with the Highest Accuracy from 10-Fold Stratified Cross-Validation. \\}
\label{tab:Best_ViTFold}
\resizebox{\columnwidth}{!}{%
\renewcommand{\arraystretch}{1.8}
 \begin{tabular}{|cc|c|c|c|c|c|}\hline 
    \multicolumn{2}{|c|}{\multirow{1}{*}{\centering \textbf{Method}}} &  
    \centering \textbf{Accuracy\%} & \centering \textbf{Sensitivity\%} & \centering 
     \textbf{Specificity\%} & \centering \textbf{Precision\%} & \centering\arraybackslash 
     \textbf{F1-score\%} \\
     \hline 
     \multicolumn{1}{|c}{\multirow{4.3}{*}{ \rotatebox[origin=c]{90}{Atlas-based models}}}& 
     \multicolumn{1}{|c}{\centering AAL-based model} & 
     \multicolumn{1}{|c}{ \begin{tabular}{c}77.73 \end{tabular}} & 
     \multicolumn{1}{|c}{\begin{tabular}{c}\textbf{92.97} \end{tabular}} & \multicolumn{1}{|c}{\begin{tabular}{c}60.00 \end{tabular}} & 
     \multicolumn{1}{|c}{\begin{tabular}{c}73.01 \end{tabular}} & 
     \multicolumn{1}{|c|}{\begin{tabular}{c}\textbf{81.79} \end{tabular}}\\ 
      \cline{2-7} 
     & \multicolumn{1}{|c}{\centering HO-based model} & \multicolumn{1}{|c}{\begin{tabular}{c}77.73 \end{tabular}} & \multicolumn{1}{|c}{\begin{tabular}{c}83.59 \end{tabular}} & \multicolumn{1}{|c}{\begin{tabular}{c}70.91\end{tabular}} & \multicolumn{1}{|c}{\begin{tabular}{c}76.98\end{tabular}} & \multicolumn{1}{|c|}{\begin{tabular}{c} 80.15 \end{tabular}}\\
     \cline{2-7} 
     &  \multicolumn{1}{|c}{\centering Dose-based model} & \multicolumn{1}{|c}{\begin{tabular}{c}\textbf{78.99}\end{tabular}} & \multicolumn{1}{|c}{\begin{tabular}{c}78.91\end{tabular}} & \multicolumn{1}{|c}{\begin{tabular}{c}\textbf{79.09}\end{tabular}} & \multicolumn{1}{|c}{\begin{tabular}{c}\textbf{81.45}\end{tabular}} & \multicolumn{1}{|c|}{\begin{tabular}{c}80.16\end{tabular}}\\ 
     \cline{2-7} 
     & \multicolumn{1}{|c}{\centering CK-based model} & \multicolumn{1}{|c}{\begin{tabular}{c}78.15 \end{tabular}} & \multicolumn{1}{|c}{\begin{tabular}{c}81.25 \end{tabular}} & \multicolumn{1}{|c}{\begin{tabular}{c}74.55\end{tabular}} & \multicolumn{1}{|c}{\begin{tabular}{c}78.79\end{tabular}} & \multicolumn{1}{|c|}{\begin{tabular}{c} 80.00\end{tabular}}\\ 
     \hline 
     \multicolumn{2}{|c}{Cube-based model}
    & \multicolumn{1}{|c}{\begin{tabular}{c}78.57\end{tabular}} & 
    \multicolumn{1}{|c}{\begin{tabular}{c}87.50\end{tabular}} & \multicolumn{1}{|c}{\begin{tabular}{c}68.18\end{tabular}} & 
    \multicolumn{1}{|c}{\begin{tabular}{c}76.19\end{tabular}}& \multicolumn{1}{|c|}{\begin{tabular}{c}81.45\end{tabular}} \\
    \hline
  \end{tabular}%
}
{\\\\The bolded values indicate the best performing results.}
\end{table}

\subsubsection{RESULTS OF 3DVIT-GAT MODELS}
This section compares the classification performance of our proposed atlas-based 3DViT-GAT models with two recent DL frameworks: Gao et al. \cite{gao2023classification} and Xiao et al. \cite{10925117}. Both models are also trained and evaluated using the REST-meta-MDD dataset.  Further, both our models and that of Gao et al. \cite{gao2023classification} utilized a stratified 10-fold cross-validation protocol to ensure a robust and generalizable performance evaluation.
Our approach is also evaluated against a cube-based baseline model that extracts regions from fixed-size volumetric partitions rather than anatomically guided partitions. The key performance metrics of all of these models are displayed in Table \ref{tab:GAT_models_performance}.

Among all evaluated models, our Dose-based 3DViT-GAT model achieved the best performance, with an accuracy of 80.88$\pm$0.39\%, a sensitivity of 86.25$\pm$2.91\%, a specificity of 74.64$\pm$3.48\%, a precision of 79.92$\pm$1.69\%, and an F1-score of 82.90$\pm$0.57\%. The AAl-based achieved the second highest performance, with an accuracy of 79.71$\pm$0.42\%, a specificity of 70.55$\pm$2.31\%, a precision of 77.61$\pm$1.04\%, and an F1-score of 82.27$\pm$0.42\%. It also exhibited a sensitivity of 87.58$\pm$1.80\%, which is higher than that of its competitors. Similarly,  the CK-based model produced 79.20$\pm$0.60\% accuracy, 83.67$\pm$1.86\% sensitivity, 74.00$\pm$2.08\% specificity, 78.95$\pm$1.05\% precision, and 81.22$\pm$0.64\% F1-score. These results demonstrate that our proposed atlas-based models provide solid and balanced classification capabilities across all evaluation metrics, emphasizing the advantages of combining atlas-guided region extraction with ViT-based representations and graph learning. In contrast, the Cube-based model reported 78.95$\pm$1.25\% accuracy, 82.03$\pm$2.99\% sensitivity, 75.36$\pm$3.87\% specificity, 79.59$\pm$2.19\% precision, and 80.73$\pm$1.20\% F1-score. However, its lack of anatomical localization may limit its ability to detect meaningful morphological patterns critical for MDD detection.

The model proposed by Xiao et al. \cite{10925117} yielded 77.17\% accuracy, 80.30\% sensitivity, 74.00\% specificity, 75.56\% precision, and 77.86\% F1-score. Although this model showed a strong performance, their Transformer-based architecture operated on uniformly segmented 3D blocks without anatomical guidance, limiting its ability to capture fine-grained brain structures relevant to accurate diagnosis. Our HO-based model also showed strong performance, with an accuracy of 77.06$\pm$0.80\%, a sensitivity of 76.09$\pm$5.45\%, a specificity of 78.18$\pm$7.14\%, a precision of 80.69$\pm$3.89\%, and an F1-score of 78.05$\pm$1.14\%. Gao et al. \cite{gao2023classification}  constructed structural covariance matrices using the AAL atlas and employed a 3D CNN to heatmap-enhanced GM images, attained an accuracy of 76.64\%, a sensitivity of 73.52\%, a specificity of 79.40\%, a precision of 88.73\%, and an F1-score of 80.41\%. Although it has high specificity and precision, its low sensitivity reflects its limited ability to detect true MDD cases, which can be costly in clinical settings. 

To support these findings, we performed two-sample t-tests (Table \ref{tab:ReslutsTTest}) comparing each atlas-based model with the Cube-based model. This study assumed that statistically significant differences existed between models when $p < 0.05$. The results showed that there was a significant difference in  sensitivity, specificity, precision, and F1-score between the AAL-based model and the Cube-based model. Meanwhile, there were significant differences between the Dose-based model and the Cube-based model in terms of in accuracy, sensitivity, and F1-score. There was also a significant difference in accuracy, sensitivity, and F1-score between the Cube-based and HO-based models, supporting the superiority of the Cube-based approach over HO. Nevertheless, the Cube-based model performed significantly worse than all other atlas-based models, highlighting the significance of structured, anatomically guided ROI extraction to improve classification performance. Additionally, there is no significant difference between the Cube-based and CK-based models.

Furthermore, Table \ref{tab:Best_ViTGATFold} illustrates the evaluation metrics for the fold with the highest accuracy for the 3DViT-GAT models. The best performing Dose-based model produced 81.51\% accuracy, 85.94\% sensitivity, 76.36\% specificity, 80.88\% precision, and 83.33\% F1-score. In AAL-based model, the best model achieved an accuracy of 80.25\%, a sensitivity of 90.63\%, a specificity of 68.18\%, a precision of 76.82\%, and an F1-score of 83.15\%. The best CK-based model had 80.25\% accuracy, 87.50\% sensitivity, 71.82\% specificity, 78.32\% precision, and 82.66\% F1-score. In addition, the best HO-based model recorded 78.15\% accuracy, 78.13\% sensitivity, 78.18\% specificity, 80.65\% precision, and 79.37\% F1-score. While, the best Cube-based model had 81.51\% accuracy, 81.25\% sensitivity, 81.82\% specificity, 83.87\% precision, and 82.54\% F1-score.

\begin{table}[htbp]
\caption{Performance Comparison of the 3DViT-GAT Models with Other Existing Models.}
\label{tab:GAT_models_performance}
\resizebox{\columnwidth}{!}{%
\renewcommand{\arraystretch}{1.8}
  \begin{tabular}{|cc|c|c|c|c|c|}\hline 
    \multicolumn{2}{|c}{\multirow{2}{*}{\textbf{Method}}} & 
    \multicolumn{5}{|c|}{ \centering \textbf{Average} $\pm$ \textbf{Standard deviation (\%)} } \\
     \cline{3-7} 
     & & \centering \textbf{Accuracy} & \centering \textbf{Sensitivity} & \centering 
     \textbf{Specificity} & \centering \textbf{Precision} & \centering\arraybackslash 
     \textbf{F1-score} \\
     \hline
    \multicolumn{2}{|c}{ Gao et al. \cite{gao2023classification}}
    & \multicolumn{1}{|c}{\begin{tabular}{c}76.64  \end{tabular}} & 
    \multicolumn{1}{|c}{\begin{tabular}{c}73.52 \end{tabular}} & \multicolumn{1}{|c}{\begin{tabular}{c}\textbf{79.40} \end{tabular}} & 
    \multicolumn{1}{|c}{\begin{tabular}{c}\textbf{88.73}  \end{tabular}}& \multicolumn{1}{|c|}{\begin{tabular}{c}80.41 \end{tabular}}
    \\
     \hline
    \multicolumn{2}{|c}{ Xaio et al.  \cite{10925117}}
    & \multicolumn{1}{|c}{\begin{tabular}{c}77.17  \end{tabular}} & 
    \multicolumn{1}{|c}{\begin{tabular}{c}80.30 \end{tabular}} & \multicolumn{1}{|c}{\begin{tabular}{c}74.00 \end{tabular}} & 
    \multicolumn{1}{|c}{\begin{tabular}{c}75.56  \end{tabular}}& \multicolumn{1}{|c|}{\begin{tabular}{c} 77.86\end{tabular}}
     
     \\
     \hline 
     \multicolumn{1}{|c}{\multirow{4.3}{*}{ \rotatebox[origin=c]{90}{Atlas-based models}}}& 
     \multicolumn{1}{|c}{\centering AAL-based model} & 
     \multicolumn{1}{|c}{ \begin{tabular}{c} 79.71$\pm$0.42\end{tabular}} & 
     \multicolumn{1}{|c}{\begin{tabular}{c}\textbf{87.58$\pm$1.80}\end{tabular}} & \multicolumn{1}{|c}{\begin{tabular}{c} 70.55$\pm$2.31 \end{tabular}} & 
     \multicolumn{1}{|c}{\begin{tabular}{c} 77.61$\pm$1.04\end{tabular}} & 
     \multicolumn{1}{|c|}{\begin{tabular}{c} 82.27$\pm$0.42\end{tabular}}\\ 
     \cline{2-7} 
     & \multicolumn{1}{|c}{\centering HO-based model} & \multicolumn{1}{|c}{\begin{tabular}{c} 77.06$\pm$0.80 \end{tabular}} & \multicolumn{1}{|c}{\begin{tabular}{c} 76.09$\pm$5.45 \end{tabular}} & \multicolumn{1}{|c}{\begin{tabular}{c}78.18$\pm$7.14 \end{tabular}} & \multicolumn{1}{|c}{\begin{tabular}{c}80.69$\pm$3.89 \end{tabular}} & \multicolumn{1}{|c|}{\begin{tabular}{c} 78.05$\pm$1.14\end{tabular}}\\
     \cline{2-7} 
     &  \multicolumn{1}{|c}{\centering Dose-based model} & \multicolumn{1}{|c}{\begin{tabular}{c} \textbf{80.88$\pm$0.39} \end{tabular}} & \multicolumn{1}{|c}{\begin{tabular}{c}86.25$\pm$2.91 \end{tabular}} & \multicolumn{1}{|c}{\begin{tabular}{c}74.64$\pm$3.48 \end{tabular}} & \multicolumn{1}{|c}{\begin{tabular}{c}79.92$\pm$1.69 \end{tabular}} & \multicolumn{1}{|c|}{\begin{tabular}{c} \textbf{82.90$\pm$0.57}\end{tabular}}\\ 
     \cline{2-7} 
     & \multicolumn{1}{|c}{\centering CK-based model} & \multicolumn{1}{|c}{\begin{tabular}{c} 79.20$\pm$0.60 \end{tabular}} & \multicolumn{1}{|c}{\begin{tabular}{c} 83.67$\pm$1.86 \end{tabular}} & \multicolumn{1}{|c}{\begin{tabular}{c}74.00$\pm$2.08 \end{tabular}} & \multicolumn{1}{|c}{\begin{tabular}{c}78.95$\pm$1.05 \end{tabular}} & \multicolumn{1}{|c|}{\begin{tabular}{c} 81.22$\pm$0.64 \end{tabular}}\\ 
     \hline 
     				
     \multicolumn{2}{|c}{Cube-based model}
    & \multicolumn{1}{|c}{\begin{tabular}{c}78.95$\pm$1.25\end{tabular}} & 
    \multicolumn{1}{|c}{\begin{tabular}{c}82.03$\pm$2.99 \end{tabular}} & \multicolumn{1}{|c}{\begin{tabular}{c}75.36$\pm$3.87 \end{tabular}} & 
    \multicolumn{1}{|c}{\begin{tabular}{c}79.59$\pm$2.19 \end{tabular}}& \multicolumn{1}{|c|}{\begin{tabular}{c} 80.73$\pm$1.20 \end{tabular}} \\
    \hline
     
  \end{tabular}%
}
{\\\\The bolded values indicate the best performing results.}
\end{table}

\begin{table}[h]
\caption{Comparison of p-values Between the 3DViT-GAT Models.\\}
\label{tab:ReslutsTTest}
\resizebox{\columnwidth}{!}{%
\renewcommand{\arraystretch}{1.6}
\begin{tabular}{|c|c|c|c|c|c|}
\hline \textbf{Compared models} & \textbf{Accuracy} & \textbf{Sensitivity} & \textbf{Specificity} & \textbf{Precision} & \textbf{F1-score} \\
\hline \begin{tabular}{c} 				
AAL-based model and Cube-based model
\end{tabular} & 0.1025& \textbf{0.0002} &\textbf{0.0048}  &  \textbf{0.0250}&\textbf{0.0018}   \\
\hline \begin{tabular}{c} 
HO-based model and Cube-based model
\end{tabular} &\textbf{0.0013}&\textbf{0.0103} & 0.3114 &  0.4687 &\textbf{0.0001}  \\
\hline \begin{tabular}{c} 
Dose-based model and Cube-based model
\end{tabular} &\textbf{0.0003}&\textbf{0.0071}  & 0.6799 & 0.7277&\textbf{0.0001}  \\
\hline \begin{tabular}{c} 
CK-based model and Cube-based model
\end{tabular} & 0.5906 & 0.17885 & 0.3637 & 0.4425 & 0.2878 \\
\hline
\end{tabular}%
}
{\\\\The bolded values indicate statistically significant differences between models at $p < 0.05$.}
\end{table}

\begin{table}[htbp]
\caption{Classification Results for the Fold with the Highest Accuracy from 10-Fold Stratified Cross-Validation. \\}
\label{tab:Best_ViTGATFold}
\resizebox{\columnwidth}{!}{%
\renewcommand{\arraystretch}{1.8}
  \begin{tabular}{|cc|c|c|c|c|c|}\hline 
    \multicolumn{2}{|c|}{\multirow{1}{*}{\centering \textbf{Method}}} &  
    \centering \textbf{Accuracy\%} & \centering \textbf{Sensitivity\%} & \centering 
     \textbf{Specificity\%} & \centering \textbf{Precision\%} & \centering\arraybackslash 
     \textbf{F1-score\%}  \\
     \hline 
     \multicolumn{1}{|c}{\multirow{4.3}{*}{ \rotatebox[origin=c]{90}{Atlas-based models}}}& 
     \multicolumn{1}{|c}{\centering AAL-based model} & 
     \multicolumn{1}{|c}{\begin{tabular}{c}80.25  \end{tabular}} & 
    \multicolumn{1}{|c}{\begin{tabular}{c} \textbf{90.63}\end{tabular}} & \multicolumn{1}{|c}{\begin{tabular}{c} 68.18\end{tabular}} & 
    \multicolumn{1}{|c}{\begin{tabular}{c} 76.82 \end{tabular}}& \multicolumn{1}{|c|}{\begin{tabular}{c} 83.15\end{tabular}} \\
      \cline{2-7} 
     & \multicolumn{1}{|c}{\centering HO-based model} & \multicolumn{1}{|c}{\begin{tabular}{c} 78.15\end{tabular}} & 
    \multicolumn{1}{|c}{\begin{tabular}{c} 78.13\end{tabular}} & \multicolumn{1}{|c}{\begin{tabular}{c} 78.18\end{tabular}} & 
    \multicolumn{1}{|c}{\begin{tabular}{c} 80.65 \end{tabular}}& \multicolumn{1}{|c|}{\begin{tabular}{c} 79.37\end{tabular}} \\
     \cline{2-7} & \multicolumn{1}{|c}{\centering Dose-based model} &
     \multicolumn{1}{|c}{\begin{tabular}{c}\textbf{81.51} \end{tabular}} & 
    \multicolumn{1}{|c}{\begin{tabular}{c} 85.94\end{tabular}} & \multicolumn{1}{|c}{\begin{tabular}{c}76.36 \end{tabular}} & 
    \multicolumn{1}{|c}{\begin{tabular}{c} 80.88 \end{tabular}}& \multicolumn{1}{|c|}{\begin{tabular}{c} \textbf{83.33}\end{tabular}} \\ 
     \cline{2-7} 
     & \multicolumn{1}{|c}{\centering CK-based model} &  \multicolumn{1}{|c}{\begin{tabular}{c} 80.25 \end{tabular}} & 
    \multicolumn{1}{|c}{\begin{tabular}{c} 87.50\end{tabular}} & \multicolumn{1}{|c}{\begin{tabular}{c} 71.82\end{tabular}} & 
    \multicolumn{1}{|c}{\begin{tabular}{c}78.32  \end{tabular}}& \multicolumn{1}{|c|}{\begin{tabular}{c} 82.66\end{tabular}} \\ 
     \hline  
     \multicolumn{2}{|c}{Cube-based model}
    & \multicolumn{1}{|c}{\begin{tabular}{c}\textbf{ 81.51} \end{tabular}} & 
    \multicolumn{1}{|c}{\begin{tabular}{c}81.25 \end{tabular}} & \multicolumn{1}{|c}{\begin{tabular}{c} \textbf{81.82}\end{tabular}} & 
    \multicolumn{1}{|c}{\begin{tabular}{c} \textbf{83.87} \end{tabular}}& \multicolumn{1}{|c|}{\begin{tabular}{c} 82.54\end{tabular}} \\
    \hline
  \end{tabular}%
}
{\\\\The bolded values indicate the best performing results.}
\end{table}

\subsubsection{Model Explainability And Interpretation}
To address our concerns about the interpretability and biological relevance of the extracted brain regions, we performed a post-hoc explainability analysis to determine which brain regions played the most significant role in the classification of MDD and HC. To achieve this, we applied a node-level interpretability approach derived from GNNExplainer \cite{ying2019gnnexplainer}. Moreover, the explainability analysis are conducted using the two best-performing atlas-based 3DViT-GAT models (Dose and AAL) in our study to ensure meaningful and representative results. For each trained model, the contribution of each node (ROI) is calculated across all subjects in the test set, separately for MDD and HC classes, as well as for the overall classification. The ROI importance scores are then averaged over all test subjects to derive reliable estimates of regional significance. We visualized the top ten ROIs on a cortical surface template using bubble plots, with bubble size and color intensity representing the importance of each ROI.

\begin{figure}[htp]
    \centering
 \begin{subfigure}{}
\includegraphics[height=0.33\textheight, width=0.9\textwidth]{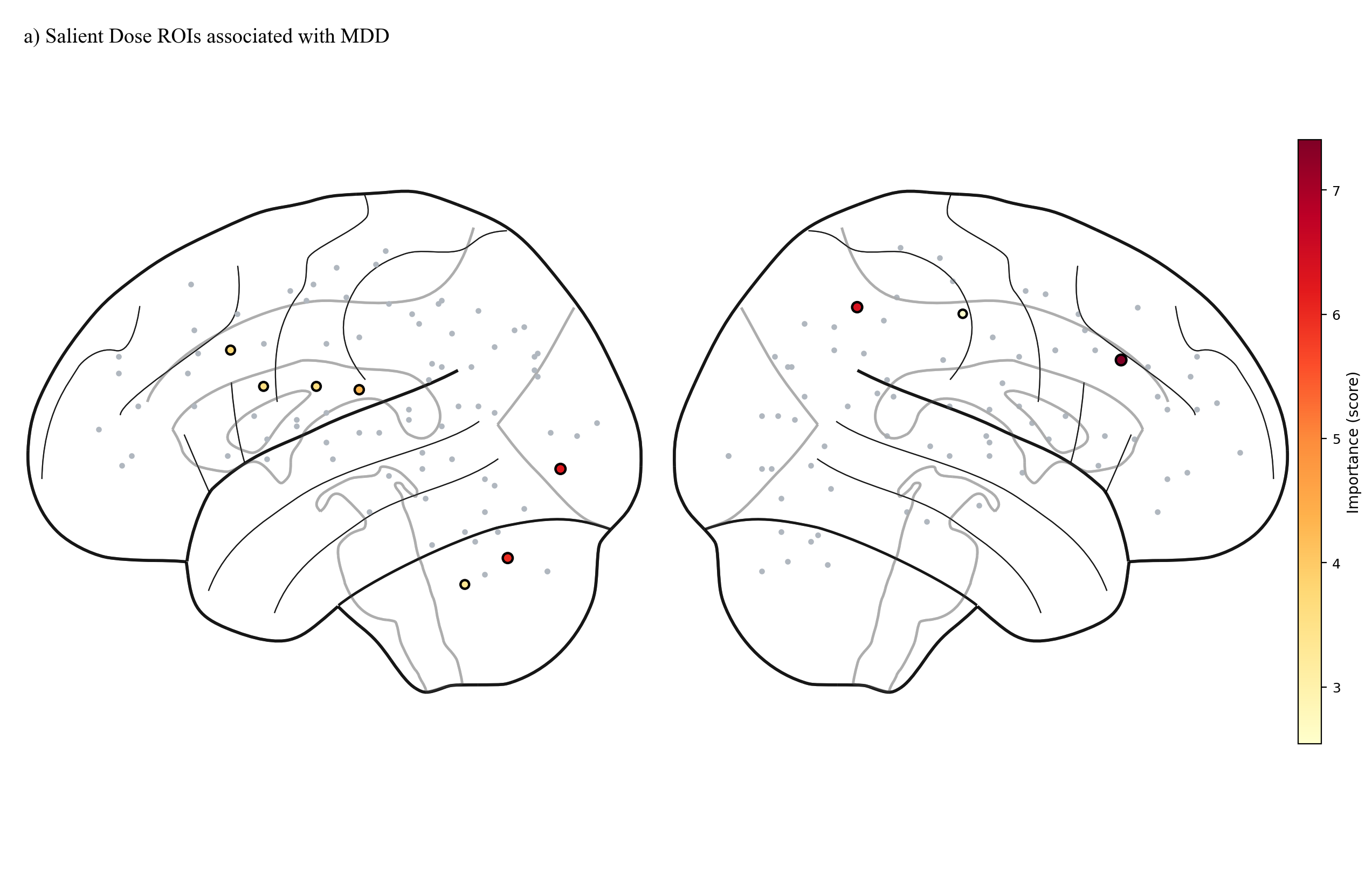}
\end{subfigure}
\vspace{-15mm}
\begin{subfigure}{}
\includegraphics[height=0.33\textheight,width=0.9\textwidth]{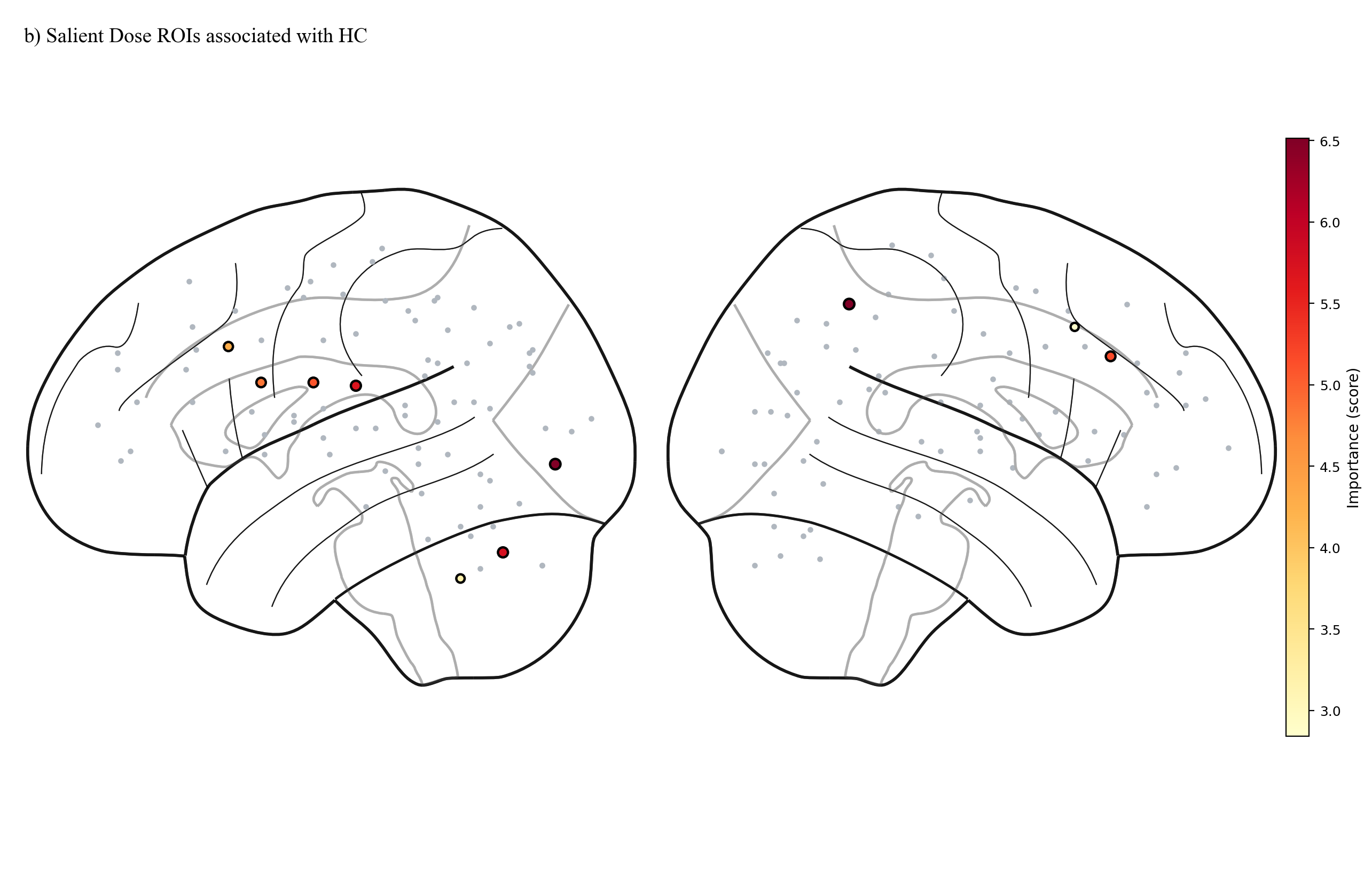}
\end{subfigure}
\vspace{-15mm}
\begin{subfigure}{}
\includegraphics[height=0.33\textheight,width=0.9\textwidth]{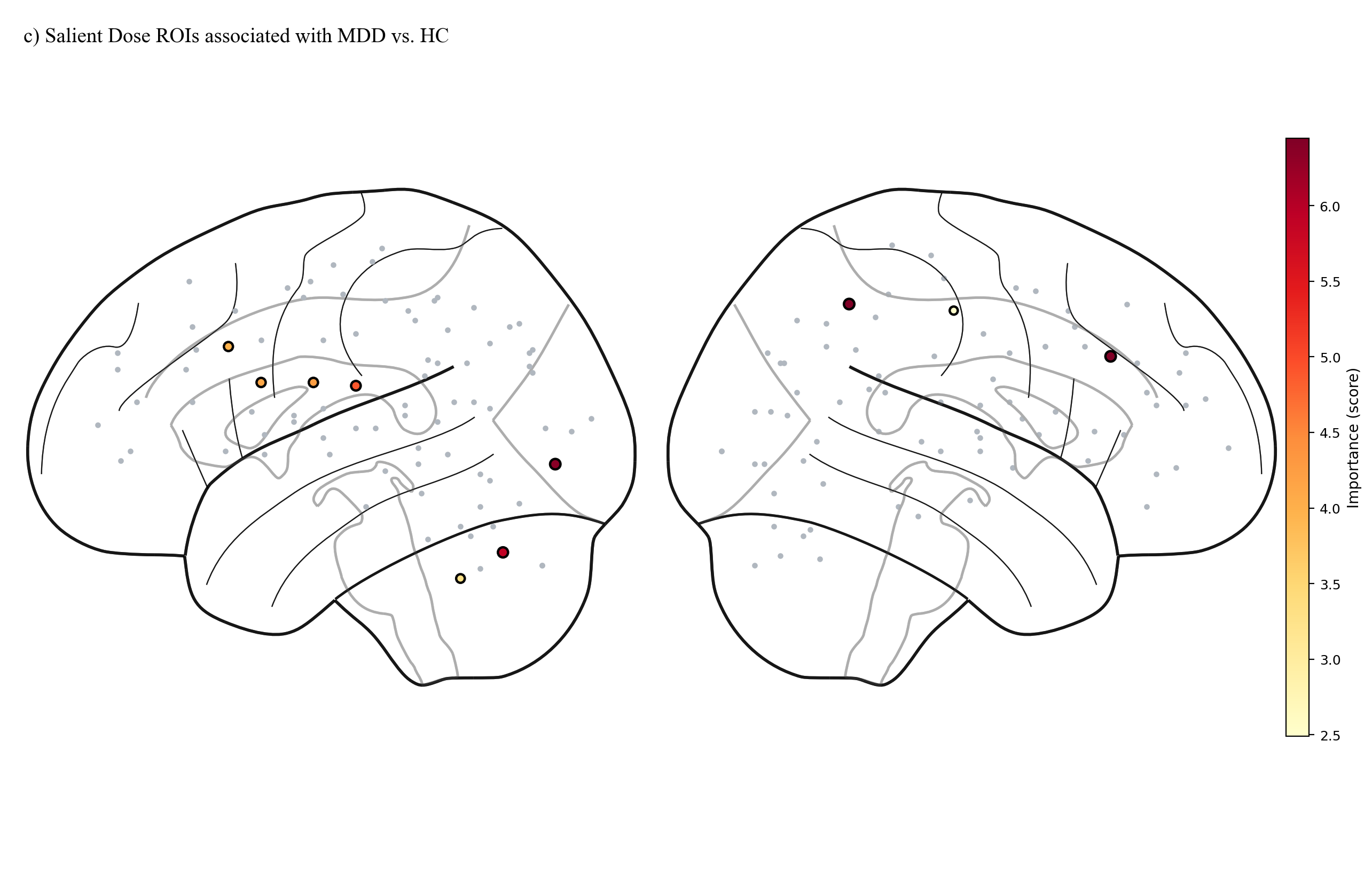}
\end{subfigure}
\caption{Explainability maps of the Dose-based model generated by GNNExplainer. a) Salient ROIs associated with MDD; b) Salient ROIs associated with HC; and c) Salient ROIs associated with MDD vs. HC. Each bubble represents an ROI, whose size and color intensity reflect its relative contribution score. Abbreviations: Dose, dosenbach atlas; ROIs, Brain regions of interest. }
\label{fig:Dos_top10_mdd_hc}
\end{figure}

Figure \ref{fig:Dos_top10_mdd_hc} (a-c) provides the explainability results derived from the Dose-based model, highlighting the most salient ROIs contributing to the classification of MDD, HC, and MDD vs. HC, respectively. All three explainability maps indicated that highly influential ROIs are located within the fronto-parietal network (FPN), occipital network (ON), cerebellum network (CN), sensorimotor network (SMN), and cingulo-opercular network (CON). A summary of the top ten salient ROIs identified within these networks across all three analyses can be found in Table \ref{tab:TopDos}.

\begin{figure}[htp]
    \centering
 \begin{subfigure}{}
\includegraphics[height=0.33\textheight, width=0.9\textwidth]{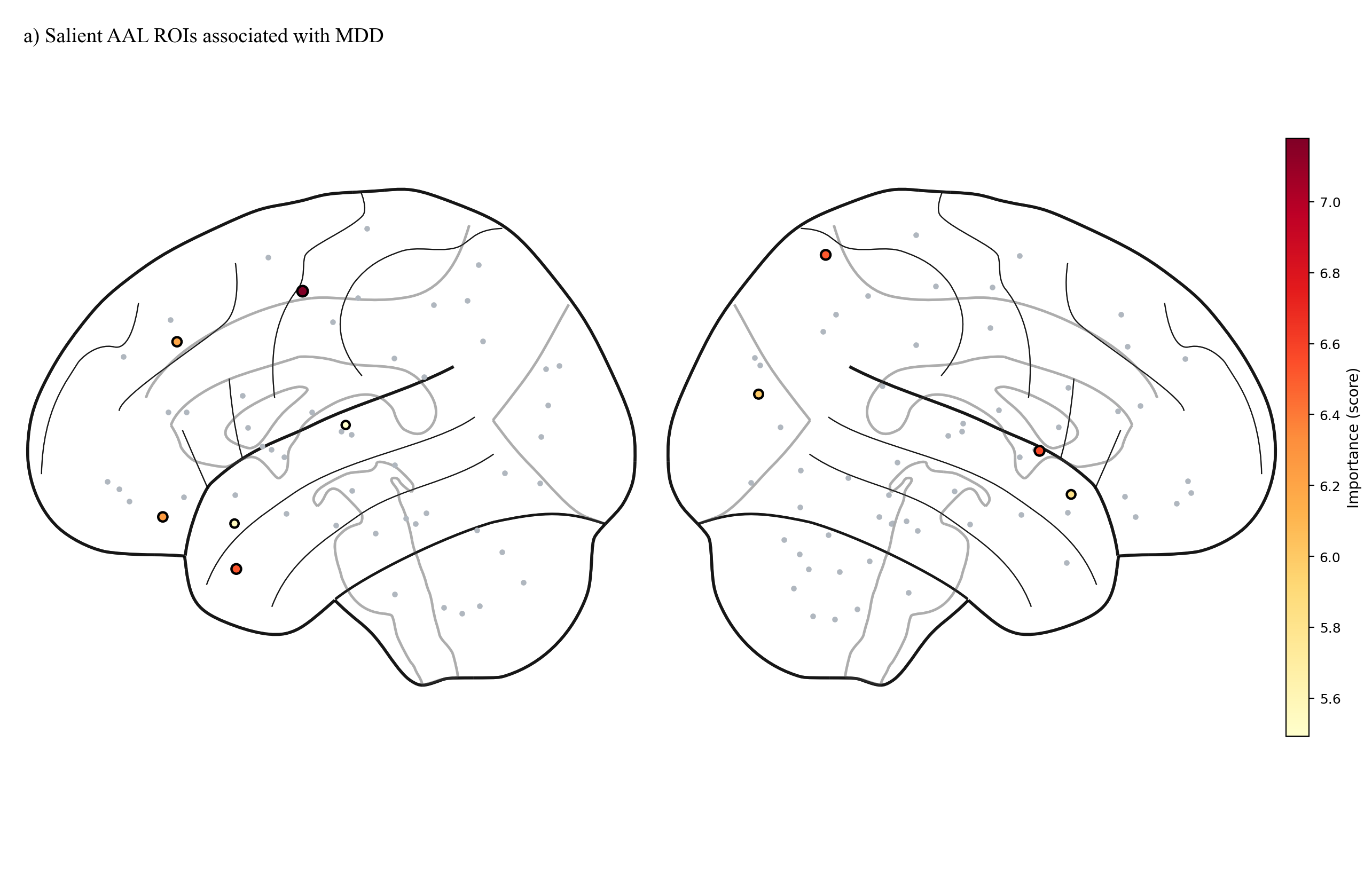}
\end{subfigure}
\vspace{-15mm}
\begin{subfigure}{}
\includegraphics[height=0.33\textheight,width=0.9\textwidth]{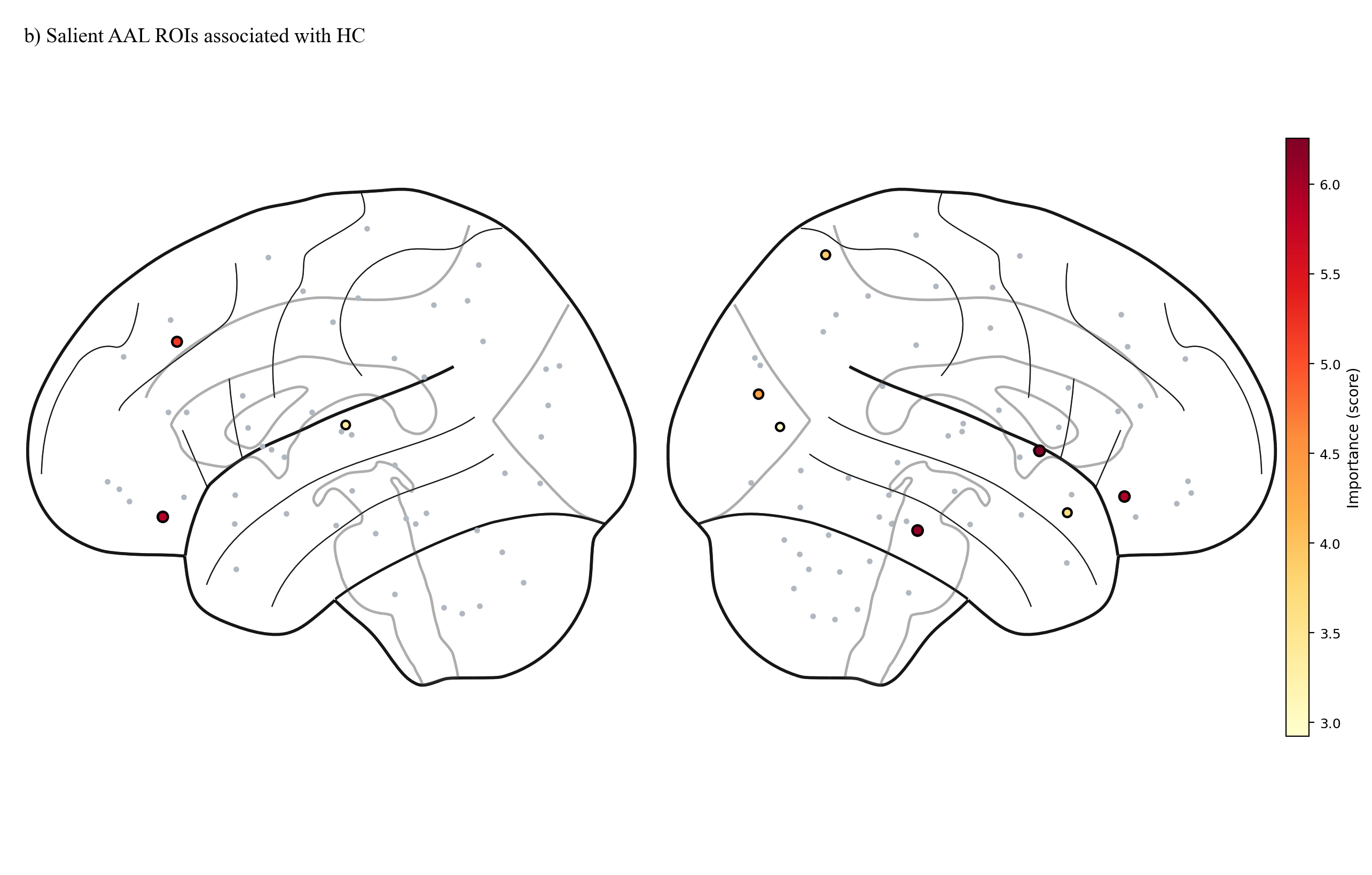}
\end{subfigure}
\vspace{-15mm}
\begin{subfigure}{}
\includegraphics[height=0.33\textheight,width=0.9\textwidth]{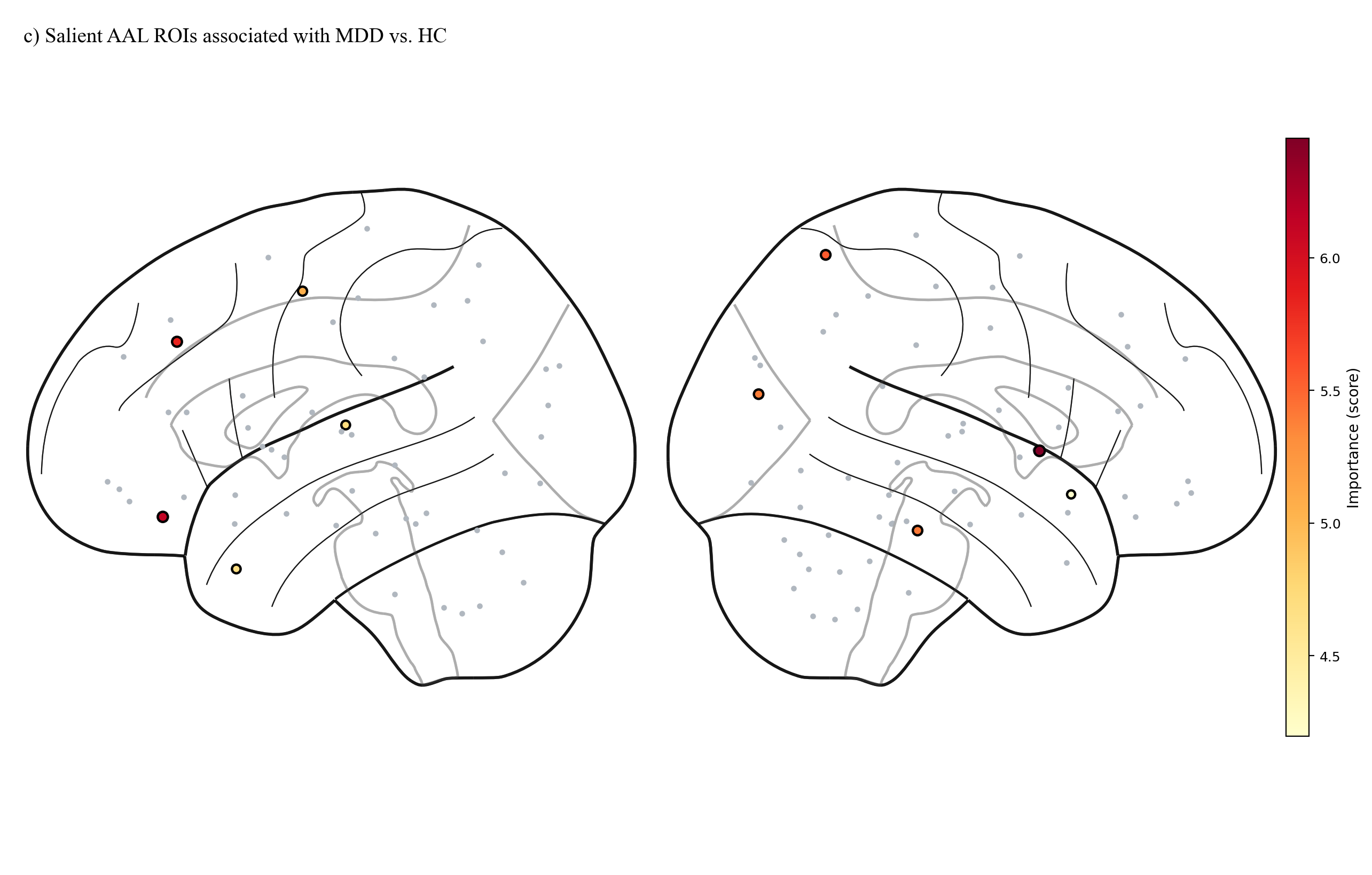}
\end{subfigure}
\caption{Explainability maps of the AAL-based model generated by GNNExplainer. a) Salient ROIs associated with MDD; b) Salient ROIs associated with HC; and c) Salient ROIs associated with MDD vs. HC. Each bubble represents an ROI, whose size and color intensity reflect its relative contribution score. Abbreviations: AAL, Automated Anatomical Labeling atlas; ROIs, Brain regions of interest.}
\label{fig:AAL_top10_mdd_hc}
\end{figure}

Subsequently, the AAL-based model is examined to identify regions that contributed to the prediction outcomes. Notably, the AAL atlas differs from the Dose atlas in that brain regions are grouped anatomically into distinct lobes rather than functional networks \cite{tzourio2002automated}. Figure \ref{fig:AAL_top10_mdd_hc} presents explainability results for the AAL-based model, emphasizing the most significant ROIs contributing to MDD, HC, and MDD vs. HC classifications. Using this model, the explainability maps revealed that the most salient ROIs are located in the frontal, insula, temporal, parietal, and occipital lobes of the brain. Table \ref{tab:TopAAL} summarizes the ten most significant ROIs within these lobes. 

\begin{table}[htp]
\caption{Top Ranked Brain Regions For Dose-based 3DViT-GAT Model.}
\label{tab:TopDos}
\resizebox{\linewidth}{!}{%
\renewcommand{\arraystretch}{1.8}
\begin{tabular}{|c|c|c|c|c|c|c|}\hline
\multirow{2}{*}{\centering  \textbf{Rank}} &
\multicolumn{2}{c|}{ \centering \small\textbf{MDD}} &
\multicolumn{2}{c|}{ \centering \textbf{HC}} &
\multicolumn{2}{c|}{ \centering  \textbf{MDD vs. HC}}\\
\cline{2-7} &
\multicolumn{1}{c|}{\textbf{Brain Region}} & \multicolumn{1}{c|}{\textbf{Network}}  &  \multicolumn{1}{c|}{\textbf{Brain Region}} & \multicolumn{1}{c|}{\textbf{Network}}  & \multicolumn{1}{c|}{\small\textbf{Brain Region}} & \multicolumn{1}{c|}{\small\textbf{Network}}  \\
\hline
1 & R dlPFC & FPN &R IPL &FPN & R IPL & FPN   \\
\hline
2 &R IPL & FPN &L post occipital& ON & R dlPFC &FPN  \\
\hline
3 &L inf cerebellum & CN &L inf cerebellum & CN& L post occipital& ON  \\
\hline
4 &L post occipital & ON &L precentral gyrus  & SMN& L inf cerebellum & CN  \\
\hline
5 &L precentral gyrus & SMN &R dlPFC & FPN &L precentral gyrus & SMN  \\
\hline
6 &L basal ganglia & CON& L precentral gyrus & SMN & L precentral gyrus & SMN \\
\hline
7 &L vFC & SMN &L vFC  & SMN &L vFC& SMN \\
\hline
8 &L precentral gyrus&  SMN &L basal ganglia  & CON&L basal ganglia &CON  \\
\hline
9 &L inf cerebellum& CN &L inf cerebellum & CN & L inf cerebellum & CN \\
\hline
10 &R parietal  & SMN &R dFC& FPN &R parietal  &SMN  \\
\hline  
\end{tabular}}
{Abbreviations: MDD, major depressive disorder; HC, healthy control; L, left hemisphere; R, right hemisphere; dlPFC, dorsolateral prefrontal cortex; FPN, fronto-parietal network; IPL, inferior parietal lobule; inf, inferior; CN, cerebellum network; ON, occipital network; SMN, sensorimotor network; CON, cingulo-opercular network; vFC, ventral frontal cortex; dFC, dorsolateral frontal cortex.}
\end{table}

\begin{table}[htp]
\caption{Top Ranked Brain Regions For AAL-based 3DViT-GAT Model.}
\label{tab:TopAAL}
\resizebox{\linewidth}{!}{%
\renewcommand{\arraystretch}{1.8}
\begin{tabular}{|c|c|c|c|c|c|c|}\hline
\multirow{2}{*}{\centering \textbf{Rank}} &
\multicolumn{2}{c|}{ \centering \textbf{MDD}} &
\multicolumn{2}{c|}{ \centering \textbf{HC}} &
\multicolumn{2}{c|}{ \centering \textbf{MDD vs. HC}}\\
\cline{2-7} &
\multicolumn{1}{c|}{\textbf{Brain Region}} & \multicolumn{1}{c|}{\textbf{Lobe}}  &  \multicolumn{1}{c|}{\textbf{Brain Region}} & \multicolumn{1}{c|}{\textbf{Lobe}}  & \multicolumn{1}{c|}{\textbf{Brain Region}} & \multicolumn{1}{c|}{\textbf{Lobe}}  \\
\hline
1 &  L PreCG & Frontal & R insula & Insula&R insula & Insula  \\
\hline
2 & R insula & Insula & R ITG &Temporal & L rectus& Frontal \\
\hline
3 & L TPOmid   & Temporal & R ORBinf &Frontal &L MFG &Frontal  \\
\hline
4 & R SPG & Parietal& L rectus &Frontal & R SPG& Parietal \\
\hline
5 & L rectus & Frontal& L MFG& Frontal&R ITG &Temporal  \\
\hline
6 & L MFG & Frontal& R MOG&Occipital &R MOG & Occipital \\
\hline
7 & R MOG&Occipital & R SPG & Parietal&L PreCG & Frontal \\
\hline
8 & R olfactory &Frontal & R TPOsup &Temporal &L heschl & Temporal \\
\hline
9 & L TPOsup &Temporal & L heschl& Temporal&L TPOmid & Temporal \\
\hline
10 & L heschl&Temporal & R  calcarine &Occipital &R olfactory & Frontal \\
\hline  
\end{tabular}}
{\\Abbreviations: MDD, major depressive disorder; HC, healthy control; L, left hemisphere; R, right hemisphere; PreCG, precentral gyrus; TPOmid, Temporal Middle Pole; SPG, Superior parietal gyrus; MFG, Middle frontal gyrus; MOG, Middle occipital gyrus; TPOsup, Temporal superior pole;  ITG, inferior temporal gyrus; ORBinf, Orbitofrontal inferior cortex.}
\end{table}

\newpage
\section{Conclusion}
In this study, we proposed a unified atlas-based framework integrating 3D ViT with GAT for MDD classification using sMRI data. This model efficiently captures intra- and inter-regional relationships by using predefined brain atlases to extract ROIs, thereby enhancing its ability to identify complex brain changes associated with MDD. Results indicated that most atlas-based models (AAL, HO, and Dose) outperformed the cube-based model, emphasizing the importance of anatomical priors for guiding the extraction of meaningful regional features. This process assisted in the precise identification of disease-relevant brain regions, improving classification performance for MDD detection.

In future work, we plan to extend the current approach by incorporating resting-state fMRI and clinical features with sMRI to build a multimodal framework. This integration is expected to provide complementary information that improves feature representations, optimizing classification accuracy and robustness. Moreover, we intend to explore ensemble strategies that combine information from multiple atlases rather than evaluating each atlas separately. Such fusion approaches could benefit from the complementary strengths of different brain parcellations and increase model generalizability.
\label{sec:conclusion}

\bibliographystyle{ieeetr}
\setcitestyle{square}
\bibliography{main}  






\end{document}